\def\year{2021}\relax
\documentclass[letterpaper]{article} 
\usepackage{aaai21}  
\usepackage{times}  
\usepackage{helvet} 
\usepackage{courier}  
\usepackage[hyphens]{url}  
\usepackage{graphicx} 
\usepackage{natbib}  
\usepackage{caption} 
\usepackage{amsmath}
\usepackage{algorithm}

\usepackage{algorithmic}
\usepackage{amsthm,amsmath,amssymb}
\usepackage{booktabs}
\usepackage{mathrsfs}

\usepackage{makecell}
\usepackage{multirow}

\begin{document}
    \title{StrokeGAN: Reducing Mode Collapse in Chinese Font Generation \\ via Stroke Encoding } 
	\author {
		Jinshan Zeng\textsuperscript{\rm 1},
		Qi Chen\textsuperscript{\rm 1},
		Yunxin Liu\textsuperscript{\rm 1},
		Mingwen Wang\textsuperscript{\rm 1}\thanks{Corresponding author},
		Yuan Yao\textsuperscript{\rm 2}$^*$\\
	}
   \affiliations {
   	\textsuperscript{\rm 1} School of Computer and Information Engineering and Institute of Artificial Intelligence, Jiangxi Normal University, Nanchang, China \\
   	\textsuperscript{\rm 2} Department of Mathematics, Hong Kong University of Science and Technology, Hong Kong \\
   	jinshanzeng@jxnu.edu.cn, chenqi970226@gmail.com, 201841600023@jxnu.edu.cn, mwwang@jxnu.edu.cn, yuany@ust.hk
   }

\maketitle

\begin{abstract}
The generation of stylish Chinese fonts is an important problem involved in many applications. Most of existing generation methods are based on the deep generative models, particularly, the generative adversarial networks (GAN) based models. However, these deep generative models may suffer from the mode collapse issue, which significantly degrades the diversity and quality of generated results. In this paper, we introduce a one-bit stroke encoding to capture the key mode information of Chinese characters and then incorporate it into CycleGAN, a popular deep generative model for Chinese font generation. As a result we propose an efficient method called \textit{StrokeGAN}, mainly motivated by the observation that the stroke encoding contains amount of mode information of Chinese characters.
In order to reconstruct the one-bit stroke encoding of the associated generated characters, we introduce a stroke-encoding reconstruction loss imposed on the discriminator.
Equipped with such one-bit stroke encoding and stroke-encoding reconstruction loss, the mode collapse issue of CycleGAN can be significantly alleviated, with an improved preservation of strokes and diversity of generated characters. The effectiveness of StrokeGAN is demonstrated by a series of generation tasks over nine datasets with different fonts. The numerical results demonstrate that StrokeGAN generally outperforms the state-of-the-art methods in terms of content and recognition accuracies, as well as certain stroke error, and also generates more realistic characters.
\end{abstract}

\section{Introduction}
\label{sc:introduction}

The stylish Chinese font generation  has attracted rising attention within recent years \cite{Lin2016,Cha2020,Chang2017,Tian2017,Kong2017,Jiang2017,Jiang2019,Chang2018,chen2019,Wu2020,Gao2020,Zhang2020}, since it has a wide range of applications including but not limited to the automatic generation of artistic Chinese calligraphy \cite{Zhao2020}, art font design \cite{Lin2014} and personalized style generation of Chinese characters \cite{Liu2012}.

The existing Chinese font generation methods can be generally divided into two categories. The first category is firstly to extract some explicit features such as strokes and radicals of Chinese characters and then utilize some traditional machine learning methods to generate new characters \cite{Xu2005,Lin2016}. The quality of feature extraction plays a central role in the first category of Chinese font generation methods. However, such feature extraction procedure is usually hand-crafted, and thus time and effort consuming.

The second category of Chinese font generation methods has been recently studied in \cite{Tian2017,Chang2018,chen2019,Gao2020,Wu2020,Zhang2020} with the development of deep learning \cite{Goodfellow2016}, particularly the generative adversarial networks (GAN) \cite{Goodfellow2014}. Due to the powerful expressivity and approximation ability of deep neural networks, feature extraction and generation procedures can be combined into one procedure, and thus, Chinese font generation methods in the second category can be usually realized in an end-to-end training way. Instead of using the stroke or radical features of Chinese characters, the methods in the second category usually regard Chinese characters directly as images, and then translate the Chinese font generation problem into certain image style translation problem \cite{Zhu2017,Isola2017}, for which GAN and its variants are the principal techniques. However, it is well-known that GAN usually suffers from the issue of mode collapse  \cite{Goodfellow2014}, that is, producing the same patterns for different inputs by generator. Such issue will significantly degrade the diversity and quality of the generated results (see, Figure \ref{fig:mode-collapse} below).
When adapted to Chinese font generation problem, the mode collapse issue will happen more frequently due to there are many Chinese characters with very similar strokes.

\begin{figure}[ht]
	\begin{minipage}[b]{0.99\linewidth}
		\centering
		\includegraphics*[scale=0.25]{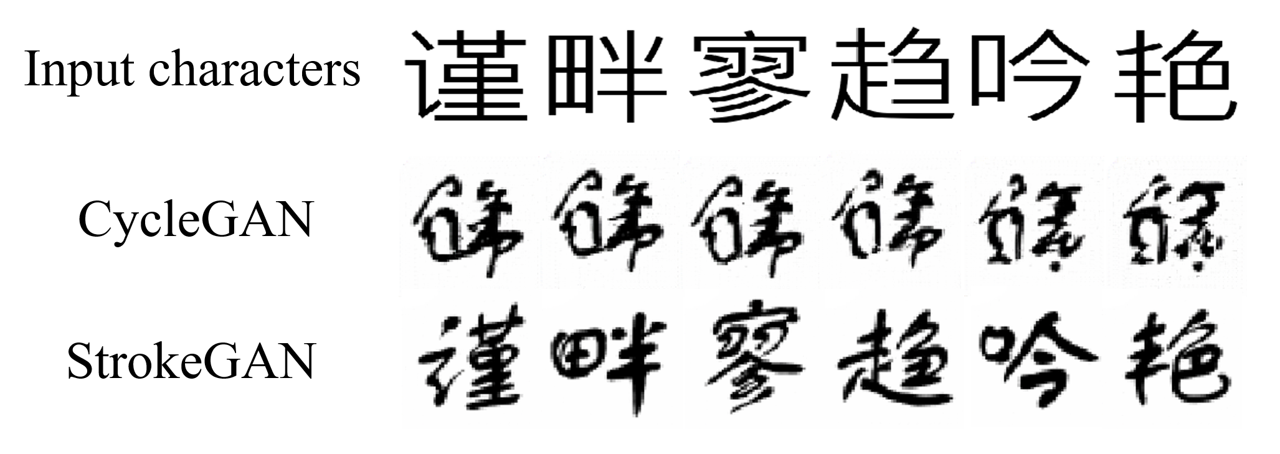}
	\end{minipage}
	\hfill
	\caption{CycleGAN for Chinese font generation suggested in \cite{Chang2018} suffers from the mode collapse issue when translating \textit{Black} font to \textit{Shu} font, while the suggested StrokeGAN can effectively tackle this issue. The settings are presented in the later experiment section.}
	\label{fig:mode-collapse}

\end{figure}

\begin{figure}[ht]
	\begin{minipage}[b]{0.99\linewidth}
		\centering
		\includegraphics*[scale=0.35]{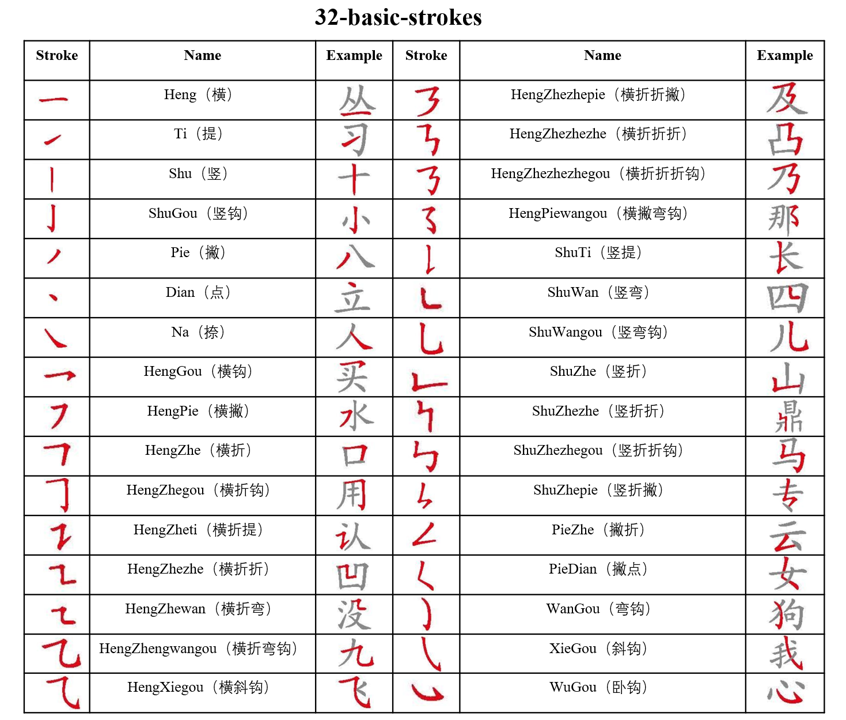}
		\centerline{{\small (a) 32 basic strokes that make up Chinese characters}}
	\end{minipage}
	\hfill
	\begin{minipage}[b]{0.99\linewidth}
		\centering
		\includegraphics*[scale=0.35]{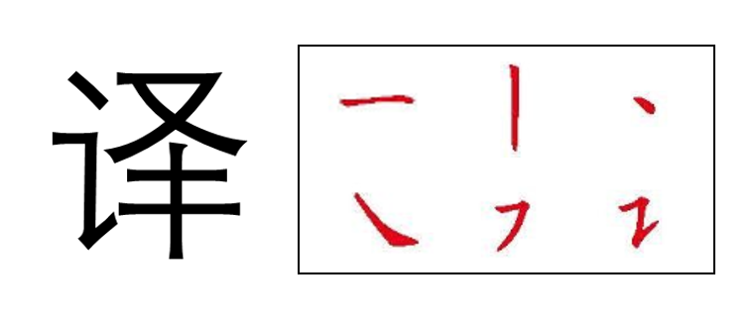}
		\centerline{{\small (b) Strokes of the Chinese character "Yi"}}
	\end{minipage}
	\hfill
	\caption{(a) 32 basic strokes that make up Chinese characters. The first and fourth columns present these 32 basic strokes, the second and fifth columns present the names of these basic strokes, and the third and sixth columns present some typical Chinese characters involved these basic strokes. (b) The strokes of the Chinese character "Yi".
	}
	\label{fig:32 basic strokes}
\end{figure}

\begin{figure}[ht]
	\begin{minipage}[b]{0.99\linewidth}
		\centering
		\includegraphics*[scale=0.35]{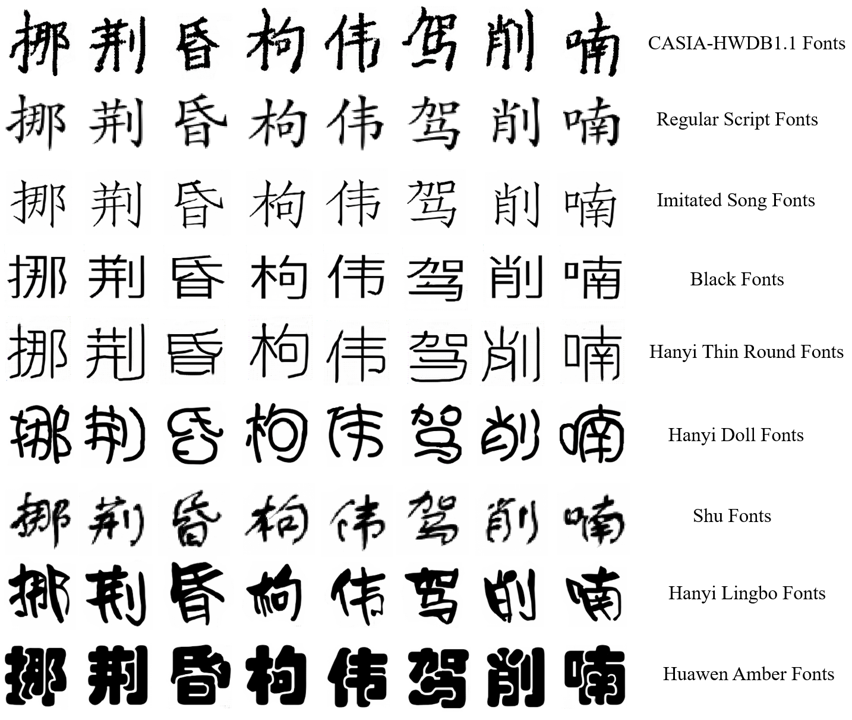}
	\end{minipage}
	\hfill
	\caption{Some Chinese characters with 9 different fonts generated by StrokeGAN. The first row presents the generated Chinese characters with the handwriting font, the second to the fourth rows present the generated Chinese characters with three different standard printing fonts, and the fifth to the ninth rows present the generated Chinese characters with five different pseudo-handwriting fonts, which are very different from the standard printing fonts and in general embody more personalized fonts.
	}
	\label{fig:stgan-results}
\end{figure}

Due to the artificial nature of Chinese characters, the explicit stroke information contains amount of mode information of Chinese characters (see Figure \ref{fig:32 basic strokes}). This is very different from natural images, which are usually regarded to be generated according to some probability distributions at some latent spaces.
Inspired by this observation, in this paper, we at first introduce a one-bit stroke encoding to preserve the key mode information of a Chinese character, and then suggest certain stroke-encoding reconstruction loss to reconstruct the stroke encoding of the generated character such that the key mode information can be well preserved, and finally incorporate them into CycleGAN for Chinese font generation \cite{Chang2018}. Thus, our suggested model is called \textit{StrokeGAN}.
The contributions of this paper can be summarized as follows:
\begin{enumerate}
\item[(a)]
We propose an effective method called \textit{StrokeGAN} for the generation of Chinese fonts with unpaired data. Our main idea is firstly to introduce a one-bit stroke encoding to capture the mode information of Chinese characters and then incorporate it into the training of CycleGAN \cite{Chang2018}, in the purpose of alleviating the mode collapse issue of CycleGAN and thus improving the diversity of its generated characters.
In order to preserve the stroke encoding, we introduce a stroke-encoding reconstruction loss to the training of CycleGAN. By the use of such one-bit stroke encoding and the associated reconstruction loss, StrokeGAN can effectively alleviate the mode collapse issue for Chinese font generation, as shown by Figure \ref{fig:mode-collapse}.

\item[(b)]
The effectiveness of StrokeGAN is verified over a set of Chinese character datasets with 9 different fonts (see Figure \ref{fig:stgan-results}), 
that is, a handwriting font, 3 standard printing fonts and 5 pseudo-handwriting fonts. Compared to CycleGAN for Chinese font generation \cite{Chang2018}, StrokeGAN can generate Chinese characters with higher quality and better diversity, particularly, the strokes are better preserved. Besides CycleGAN \cite{Chang2018}, our method also outperforms the other state-of-the-art methods including zi2zi \cite{Tian2017} and Chinese typography transfer (CTT) method \cite{Chang2017} using paired data, in terms of generating Chinese characters with higher quality and accuracy. Some generated characters by our method with 9 different fonts can be found in Figure \ref{fig:stgan-results}. It can be observed that these generated characters of StrokeGAN are very realistic.
\end{enumerate}

\subsection{Related work}

In recent years, many generation methods of stylish Chinese fonts have been suggested in the literature \cite{Tian2017,Chang2017,Chang2018,chen2019,Wu2020,Jiang2017,Jiang2019,Zhang2020} with the development of deep learning. In \cite{Tian2017}, the authors adapted \textit{pix2pix} model developed in \cite{Isola2017} for the image style translation problem to Chinese font generation and then suggested \textit{zi2zi} method with paired training data, that is, there is a one-to-one correspondence between the characters in the source (input) style domain and target (output) style domain. Similar idea was extended to realize the Chinese character generation from one font to multiple fonts in \cite{chen2019}. Besides \cite{Tian2017} and \cite{chen2019}, some other paired data based Chinese font generation methods were suggested in \cite{Chang2017,Jiang2017,Wu2020}. However, it is usually human-intensive to build up the paired training data. In order to overcome this challenge, \cite{Chang2018} adapted CycleGAN developed in \cite{Zhu2017} for the image style translation to Chinese font generation based on unpaired training data. Yet, the CycleGAN based method suggested in \cite{Chang2018} (called \textit{CCG-CycleGAN}) may suffer from the mode collapse issue \cite{Goodfellow2014}. When mode collapse occurs, the generator produces fewer patterns for different inputs, and thus significantly degrades the diversity and quality of generated results.

Motivated by the observation from traditional Chinese character generation and recognition methods (see \cite{Xu2005,Kim-stroke1999}) that the explicit stroke feature can provide much mode information for a Chinese character, in this paper, we incorporate such stroke information into the training of CycleGAN for Chinese font generation \cite{Chang2018} to tackle the issue of mode collapse, via introducing a one-bit stroke encoding and certain stroke-encoding reconstruction loss.
The very recent papers \cite{Wu2020,Jiang2019,Zhang2020} also incorporated some stroke or radical information of Chinese characters into Chinese font generation. Their main idea is firstly to utilize a deep neural network to extract the strokes or radicals of Chinese characters and then merge them by another deep neural network, which is very different to our idea of the use of a very simple one-bit stroke encoding. According to our later numerical experiments, our introduced one-bit stroke encoding is very effective.



The rest of this paper is organized as follows. In Section 2\ref{sc:preliminary work}, we present some preliminary work. In Section 3\ref{sc:Stroke GAN}, we introduce the proposed method in detail. In Section 4\ref{sc:experiments}, we provide a series of experiments to demonstrate the effectiveness of the proposed method. We conclude this paper in Section 5\ref{sc:conclusion}.

\begin{figure*}[ht]
	\begin{minipage}[b]{1\linewidth}
		\centering
		\includegraphics*[scale=0.43]{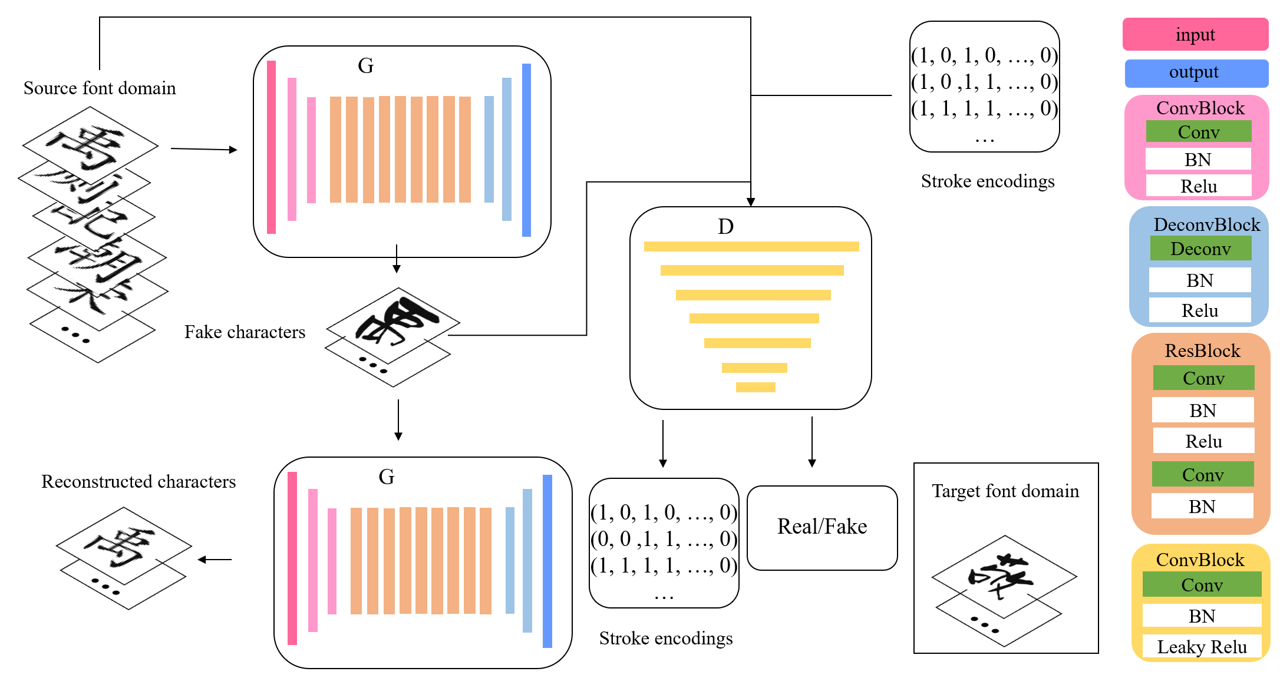}
	\end{minipage}
	\hfill
	\caption{The diagram of StrokeGAN for Chinese font generation. There are a discriminator $D$ and a generator $G$ in StrokeGAN. The workflow of StrokeGAN can be divided into three parts: (a) we at first yield the stroke encodings of Chinese characters via certain one-bit encoding way, and then send the characters to $G$ as input; (b) $G$ at first generates a fake character according to the input character from the source font domain, and later reconstructs a character in the source font domain from the generated fake character in the target font domain such that the reconstructed character is as similar as possible to the original input character; (c) $D$ takes the fake character generated by $G$ and the real character in the source font domain as the input, and attempts to on one hand distinguish whether the generated character is real or fake and on the other hand reconstruct the stroke encoding of the generated character as accurately as possible, compare with the real stroke encoding to preserve the mode information of Chinese characters. The network architectures of generator and discriminator are stacked by the modules presented in the right column.
	}
	\label{fig:stgan-model}
\end{figure*}

\section{Preliminary Work}
\label{sc:preliminary work}

In this section, we introduce some preliminary work, which serves as the basis of this paper.

\subsection{Generative Adversarial Networks}
\label{sc:Generative adversarial networks}
Generative adversarial networks (GAN) \cite{Goodfellow2014} have achieved great achievements in the task of high-quality image synthesis. The classic GAN model consists of two parts: a generator $G$ and a discriminator $D$. The generator $G$ generates fake images, and the discriminator $D$ judges whether the images generated by $G$ are fake or real. Mathematically, GAN can be formulated as the following two-player minimax game between $G$ and $D$,
\begin{align*}
\min_G \max_D \ \mathbb{E}_{x\sim \mathbb{P}_{data}} [\log D(x)] + \mathbb{E}_{z\sim \mathbb{P}_{z}} [\log(1-D(G(z)))]
\end{align*}
where $\mathbb{P}_{data}$ and $\mathbb{P}_{z}$ are the distributions of data $x$ and the input noise variable $z$ for the generator, respectively. In practice, the generator $G$ and discriminator $D$ are generally represented by some deep neural networks.

\subsection{Conditional GAN}
\label{sc:Conditional-GAN-(cGAN)-and-Cycle-GAN-(CycleGAN)}

The conditional GAN (cGAN) was suggested in \cite{Mirza2014} mainly to embed some conditional information such as the category information of samples. Such idea was later exploited in \cite{Isola2017} for the image style translation. For cGAN, besides the original input $z$, the conditional information $c$ is also a part of input for $G$. Thus, the objective for GAN should be slightly modified for cGAN, shown as follows:
\begin{align*}
{\cal L}_{adv}(D,G)
&=\mathbb{E}_{x\sim \mathbb{P}_{data}} [\log D(x)] \\
&+ \mathbb{E}_{z\sim \mathbb{P}_{z}, c\sim \mathbb{P}_c} [\log(1-D(G(z,c)))], \nonumber
\end{align*}
where $\mathbb{P}_c$ represents the distribution of the referred conditional information $c$.

\subsection{CycleGAN}
The training of cGAN \cite{Isola2017} is based on the paired data, of which the collection is usually time-consuming and laborious. In order to overcome this challenge, CycleGAN was proposed in \cite{Zhu2017} for the image style translation based on the unpaired data. The main idea of CycleGAN is to preserve key attributes between the source and target domains by utilizing a cycle consistency loss. Specifically, let $x\in {\cal X}$ and $x'\in {\cal X}'$ be two images from two different style domains conditioned on respectively some conditional information domains $c\in {\cal C}$ and $c' \in {\cal C}'$. In order to realize the bidirectional translation between them, two generators $G:{\cal X} \times {\cal C} \rightarrow {\cal X}'$ and $G':{\cal X}' \times {\cal C}' \rightarrow {\cal X}$ are exploited. With these, the cycle consistency loss can be defined as follows:
\begin{align*}
{\cal L}_{cyc}(G,G')
&= \mathbb{E}_{x\sim \mathbb{P}_x, c \sim \mathbb{P}_c, c' \sim \mathbb{P}_{c'}}[\|x-G'(G(x,c),c')\|_1]\\
&+\mathbb{E}_{x'\sim \mathbb{P}_{x'}, c' \sim \mathbb{P}_{c'}, c \sim \mathbb{P}_c}[\|x'-G(G'(x',c'),c)\|_1],
\end{align*}
where $x\in {\cal X}$, $x'\in {\cal X}'$, $c\in {\cal C}$, $c'\in {\cal C}'$, and $\mathbb{P}_*$ represents the distribution of the associated domain $*$. The generators $G$ and $G'$ are trained to make the cycle consistency loss ${\cal L}_{cyc}(G,G')$ small.

\section{StrokeGAN for Chinese Font Generation}
\label{sc:Stroke GAN}

In this section, we describe the proposed \textit{StrokeGAN} for automatic generation of stylish Chinese font. The core idea of StrokeGAN is to incorporate some one-bit stroke encodings of Chinese characters into CycleGAN to alleviate the issue of mode collapse, as presented in Figure \ref{fig:stgan-model}, mainly motivated by the basic observation that the stroke information embodies amount of mode information of Chinese characters. Recent studies suggested that mode collapse in GANs might be a by-product of removing sparse outlying modes toward robust estimation \cite{Yao2019iclr,Yao2020jmlr}. So a natural strategy to alleviate mode collapse is to enforce faithful conditional distribution reconstruction, conditioning on important modes represented by stroke encoding here. From Figure \ref{fig:stgan-model}, we at first yield the stroke encodings of Chinese characters via certain one-bit encoding way, and then take the characters in the source font domain as the inputs of generator $G$. After $G$, we yield a fake character in the target domain, then send such fake character to both generator and discriminator, where generator tries to reconstruct the character in the source font domain, and discriminator attempts to distinguish whether the generated character is real or fake and also reconstruct its stroke encoding. Thus, distinguished with the original CycleGAN for Chinese font generation in \cite{Chang2018}, there are two parts in the discriminator $D$, i.e.,
\[
D : x \rightarrow (D_{src}(x), D_{st}(x)),
\]
where $D_{src}(x)$ and $D_{st}(x)$ represent respectively the probability distribution over the source font domain and its stroke encoding for a given character $x$.

\subsection{Stroke Encoding}
\label{sc:structure loss}

From Figure \ref{fig:stgan-model}, the stroke encoding of character is taken as a part of input of CycleGAN. To realize this, we introduce a simple one-bit encoding way to yield the associated stroke encoding for a given Chinese character. Specifically, according to Figure \ref{fig:32 basic strokes}, there are in total 32 kinds of strokes to make up Chinese characters. Thus, for any given Chinese character $x$, we define its stroke encoding as a 32-dimensional vector $c \in \{0,1\}^{32}$ with the $i$-th entry being $1$ if the $i$-th kind of stroke is included in $x$ and otherwise $0$ for all $i$ from $1$ to $32$. In this paper, we only use the indicator function of such kind of stroke instead of its exact number for a given Chinese character mainly in consideration of the robustness of StrokeGAN. This is in general sufficient according to our later experiments (see, Figure \ref{fig:comp-similar-character}).

\subsection{Training Loss for StrokGAN}
\label{sc:Structure GAN for CCG}

The training loss for StrokeGAN consists of three parts, that is, the general adversarial loss, cycle consistency loss and stroke-encoding reconstruction loss, where the \textit{stroke-encoding reconstruction loss} is firstly introduced in this paper for the generation of stylish Chinese fonts.

\begin{table*}
	\caption{The performance of StrokeGAN in 9 generation tasks via comparing with CycleGAN \cite{Chang2018}. }
	\label{tab:result-number}
	\begin{center}
		\begin{tabular}{|l|c|c|c|c|c|c|c|}\hline
			
			\multirow{2}*{Character style translation} &
			\multicolumn{2}{c|}{Content accuracy (\%) $\uparrow$} & \multicolumn{2}{c|}{Recognition accuracy (\%) $\uparrow$} & \multicolumn{2}{c|}{Stroke error ($\times 10^{-2}$) $\downarrow$} \\
			\cline{2-3} \cline{4-5} \cline{6-7}
			& CycleGAN & StrokeGAN & CycleGAN & StrokeGAN & CycleGAN & StrokeGAN\\ \hline
			\textit{Regular Script}  $\rightarrow$ \textit{Shu}          &  89.56         & \textbf{90.48} &  89.36          & \textbf{90.52}  &6.79     & \textbf{5.69}\\\hline
			\textit{Regular Script}  $\rightarrow$ \textit{Huawen Amber} &  86.88         & \textbf{88.68} &  87.56          & \textbf{88.92}  &8.71     &\textbf{7.20}\\\hline
			\textit{Regular Script}  $\rightarrow$ \textit{Hanyi Lingbo} &  87.64         & \textbf{88.24} &  87.48          & \textbf{88.32}  &7.90     &\textbf{6.81}\\\hline
			\textit{Regular Script}  $\rightarrow$ \textit{Imitated Song}&  90.28         & \textbf{91.72} &  90.84          & \textbf{91.60}  &7.33     &\textbf{5.55} \\\hline
			\textit{Regular Script}  $\rightarrow$ \textit{Handwriting}  &  87.08         & \textbf{87.64} &  87.12          & \textbf{87.60}  &7.67     &\textbf{6.50}\\\hline
			\textit{Hanyi Thin Round}  $\rightarrow$ \textit{Hanyi Doll} &  87.00         & \textbf{87.60} &  86.92          & \textbf{87.80}  &7.69     &\textbf{6.52} \\\hline
			\textit{Black}  $\rightarrow$ \textit{Hanyi Thin Round}      &  86.60         & \textbf{87.76} &  86.60          & \textbf{87.92}  &7.79     &\textbf{6.96}\\\hline
			\textit{Imitated Song} $\rightarrow$ \textit{Black}          &  \textbf{89.56}& 88.96          &  \textbf{89.44} & 89.12           &8.06     & \textbf{6.72} \\\hline
			\textit{Imitated Song} $\rightarrow$ \textit{Regular Script} &  89.96         & \textbf{90.32} &  90.04          & \textbf{90.28}  &8.49     &\textbf{6.28} \\
			\hline
		\end{tabular}
	\end{center}
\end{table*}

{\bf A. Adversarial loss.} The first part of loss is the adversarial loss defined commonly as follows,
\begin{align}
\label{Eq:ad-loss}
{\cal L}_{adv}(D,G)
&= \mathbb{E}_{x}[\log D_{src}(x)] \\
&+ \mathbb{E}_{x}[\log(1-D_{src}(G(x)))], \nonumber
\end{align}
where the generator $G$ generates the fake character $G(x)$ conditional over the input character $x$, and the first part of discriminator $D_{src}$ attempts to distinguish the generated character is real or fake.

{\bf B. Cycle consistency loss.}
The second part of loss is the cycle consistency loss, which is introduced to let the generator reconstruct the character in the source font domain from the generated fake character, and thus avoid using the paired data. Specifically, such part of loss can be defined as follows:
\begin{align}
\label{Eq:cycle-consistency-loss}
{\cal L}_{cyc}(G)
=\mathbb{E}_{x}[\|x-G(G(x))\|_1],
\end{align}
where $G(x)$ represents the generated fake character according to the input pair $x$ and $G(G(x))$ represents the reconstructed character from the character $G(x)$ generated by $G$.

{\bf C. Stroke-encoding reconstruction loss.}
Notice that in both adversarial loss and cycle consistency loss, the characters are regarded as images during the training, while the stroke information is not paid much attention. Actually, as discussed before, the stroke information embodies amount of mode information of a Chinese character, and thus is some kind of very important information that should be preserved during the training. Also, the stroke information is very special for Chinese characters and makes the generation of Chinese characters very different from the style translation of natural images. Specifically, the \textit{stroke-encoding reconstruction loss} can be defined as follows:
\begin{align}
\label{Eq:structural-information-loss}
{\cal L}_{st}(D) = \mathbb{E}_{x,c} \left[{\|D_{st}(G(x))-c\|_2} \right].
\end{align}
where $D_{st}(G(x))$ is the stroke encoding yielded by discriminator for the generated fake character $G(x)$. Such stroke-encoding reconstruction loss is used to guide the networks to reconstruct the stroke encodings as accurately as possible so that the modes of characters can be preserved much better.

{\bf D. Total training loss.}
Combining the above three parts of loss, i.e., \eqref{Eq:ad-loss}-\eqref{Eq:structural-information-loss}, the total training loss of StrokeGAN is presented as follows:
\begin{align}
\label{Eq:total-loss}
{\cal L}_{strokegan}(D,G)
&= {\cal L}_{adv}(D,G) +\lambda_{cyc} {\cal L}_{cyc}(G) \\
&+ \lambda_{st}{\cal L}_{st}(D),\nonumber
\end{align}
where $\lambda_{cyc}$ and $\lambda_{st}$ are two penalty parameters. Based on the above defined loss ${\cal L}_{strokegan}(D,G)$, the discriminator $D$ attempts to maximize it while the generator $G$ tries to minimize it, shown as follows:
\begin{align}
\label{Eq:stgan-training}
\min_G \max_D \ {\cal L}_{strokegan}(D,G).
\end{align}

\section{Numerical Experiments}
\label{sc:experiments}
In this section, we provide a series of experiments to demonstrate the effectiveness of the suggested StrokeGAN. All experiments were carried out in Pytorch environment running Linux, AMD(R) Ryzen 7 2700x eight-core processor $\times16$ CPU, GeForce RTX 2080 GPU. Our codes are available in \url{https://github.com/JinshanZeng/StrokeGAN}.

\subsection{Experiment Settings}
\label{sc:exp-settings}

\begin{figure}[ht]
	\begin{minipage}[b]{0.99\linewidth}
		\centering
		\includegraphics*[scale=0.24]{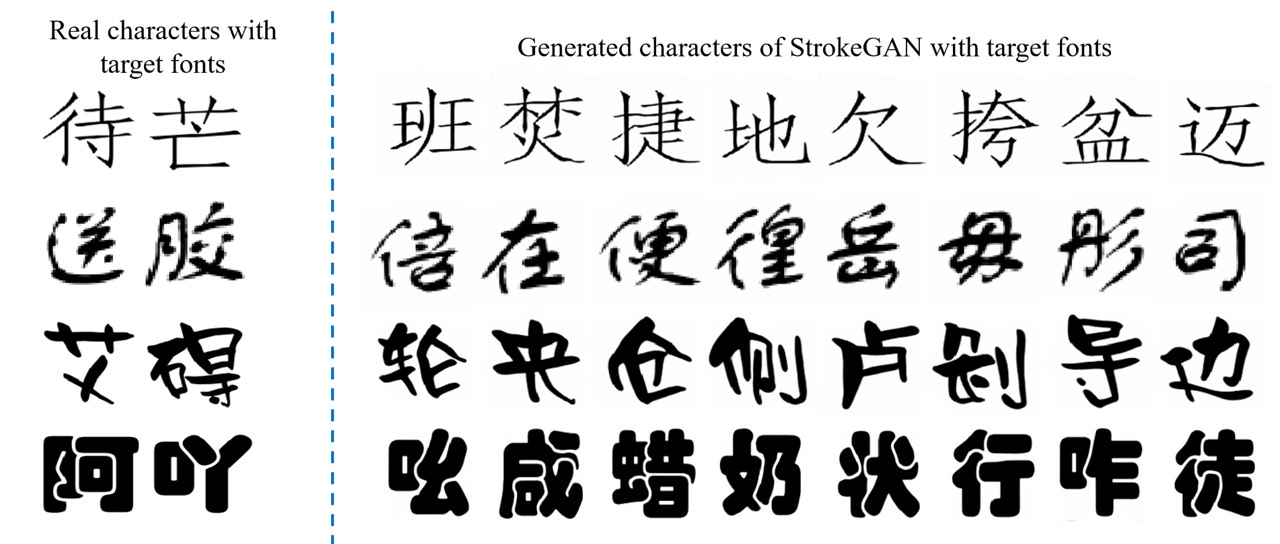}
	\end{minipage}
	\hfill
	\caption{Examples of Chinese characters generated by StrokeGAN. The characters in the left-hand side are real characters in the target font domains, while the characters in the right-hand side are generated characters in the associated target font domains. From the first row to the fourth row, the target fonts are $\{$\textit{Imitated Song}, \textit{Shu}, \textit{Hanyi Lingbo}, \textit{Huawen Amber}$\}$, respectively. It can be observed that the characters generated by StrokeGAN are very realistic.
	}
	\label{fig:results-4styles}
\end{figure}

\begin{figure}[ht]
	\begin{minipage}[b]{0.99\linewidth}
		\centering
		\includegraphics*[scale=0.21]{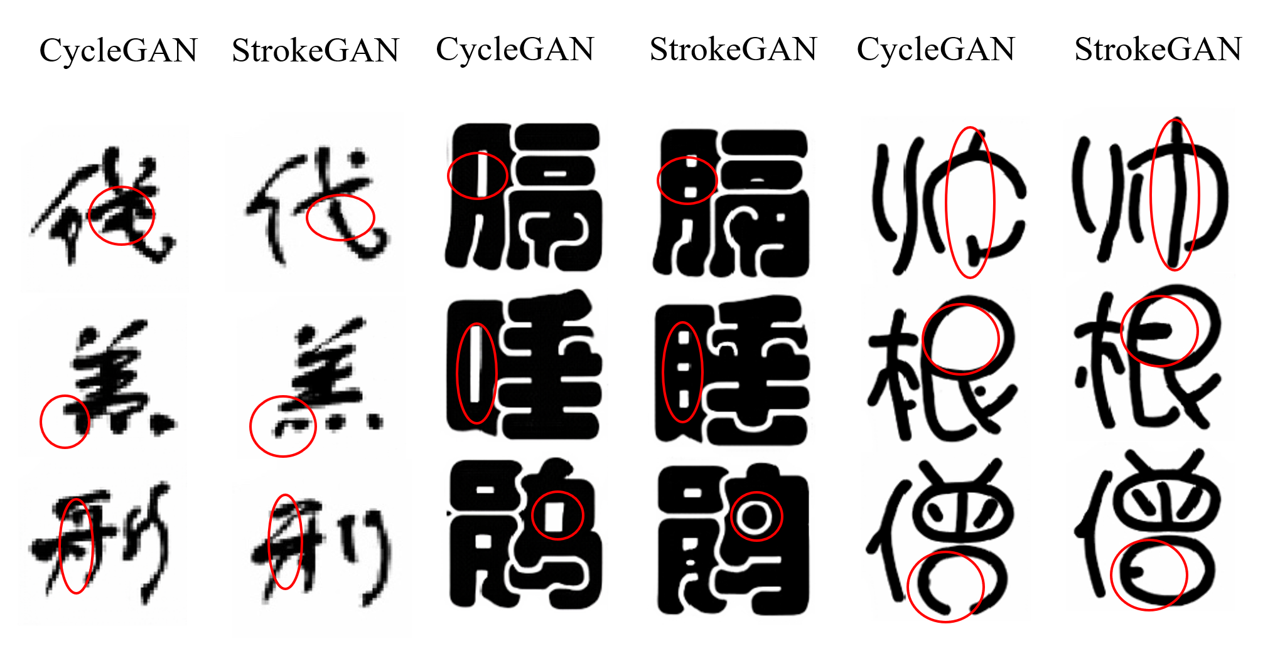}
	\end{minipage}
	\hfill
	\caption{Stoke-missing phenomena in the generated characters of CycleGAN \cite{Chang2018}. The suggested StrokeGAN can preserve strokes much better. The first two columns, the third and fourth columns, and the last two columns respectively present the generated characters from \textit{Regular Script} to \textit{Shu}, from \textit{Regular Script} to \textit{Huawen Amber}, and from \textit{Hanyi Thin Round} to \textit{Hanyi Doll}.
	}
	\label{fig:st-cycle}
\end{figure}

{\bf A. Collection of datasets.}
The dataset used in this paper consists of 9 sub-datasets with different fonts divided into three categories, i.e., a handwriting font, three standard printing fonts $\{$\textit{Black}, \textit{Regular Script}, \textit{Imitated Song}$\}$, and five pseudo-handwriting fonts $\{$\textit{Huawen Amber}, \textit{Shu}, \textit{Hanyi Lingbo}, \textit{Hanyi Doll}, \textit{Hanyi Thin Round}$\}$, where the pseudo-handwriting fonts are personalized fonts designed by artistic font designers.

The first kind of dataset related to the handwriting Chinese characters is built up from \textit{CASIA-HWDB1.1} \footnote{\url{http://www.nlpr.ia.ac.cn/databases/handwriting/Home.html}}, which was collected by 300 people.
There are in total 3755 different commonly used Chinese characters written by everyone, and thus there are $3755 \times 300$ handwriting characters in this dataset.
To build up our handwriting dataset, for each character, we randomly selected one sample from these 300 samples. Therefore, the size of the handwriting font dataset used in this paper is 3755. Except the handwriting Chinese characters, the other font datasets were collected by ourselves from the internet \footnote{say, \url{http://fonts.mobanwang.com/}} and made automatically via TTF. Specifically, the second kind of datasets contain three printing font datasets, of which the sizes are 2560, 3757 and 2506 respectively for the \textit{Black}, \textit{Regular Script} and \textit{Imitated Song} fonts. The third kind of datasets consist of five pseudo-handwriting fonts, of which the sizes are 3596, 2595, 3673, 3213 and 2840 respectively for the  \textit{Huawen Amber}, \textit{Shu}, \textit{Hanyi Lingbo}, \textit{Hanyi Doll} and \textit{Hanyi Thin Round} fonts. The size of each character was resized to $128 \times 128 \times 3$. In our experiments, we used 90\% and 10\% of the samples respectively as the training and test sets.

{\bf B. Network architectures and optimizer.}
The network structure of the generator in StrokeGAN is the same as CycleGAN \cite{Zhu2017}, including 2 convolutional layers in the down-sampling module, 9 residual modules with 2 convolutional layers of residual networks for each residual module and 2 deconvolutional layers in the up-sampling module, as presented in Table \ref{tab:generator-structure} in Appendix. The network structure of the discriminator in StrokeGAN is similar to PatchGAN \cite{Isola2017} with 6 hidden convolutional layers and 2 convolutional layers in the output module, as presented in Table \ref{tab:discriminator-structure} in Appendix. Moreover, the batch normalization \cite{Ioffe2015} was used in all layers.

In our experiments, we used the popular Adam algorithm \cite{Kingma2014} as the optimizer with the associated parameters $(0.5, 0.999)$ in both the generator and discriminator optimization subproblems. The penalty parameters of the cycle consistency loss and stroke reconstruction loss were fine-tuned at 10 and 0.18, respectively.

{\bf C. Evaluation metrics.}
To evaluate the performance of StrokeGAN, we introduced three evaluation metrics. The first one is the commonly used \textit{content accuracy} \cite{Zhu2017}, which was suggested to justify the quality of the contents of generated characters. Specifically, a pre-trained HCCG-GoogLeNet  \cite{Szegedy2014} was exploited to calculate the content accuracy. Besides the \textit{content accuracy}, we also suggested the \textit{recognition accuracy} and \textit{stroke error} particularly for the Chinese character generation task. The \textit{recognition accuracy} is defined as the ratio of those generated characters that can be correctly recognized by people to all the generated characters for testing, via a crowdsourcing way. Specifically, we randomly invited five Chinese adults to recognize the generated Chinese characters, and then took their average as the \textit{recognition accuracy}. The \textit{stroke error} is defined as the ratio of the number of missing and redundant strokes in the generated Chinese characters to its true total number of strokes. Thus, the smaller stroke error means the strokes are preserved better.

\begin{figure}[ht]
	\begin{minipage}[b]{0.99\linewidth}
		\centering
		\includegraphics*[scale=0.27]{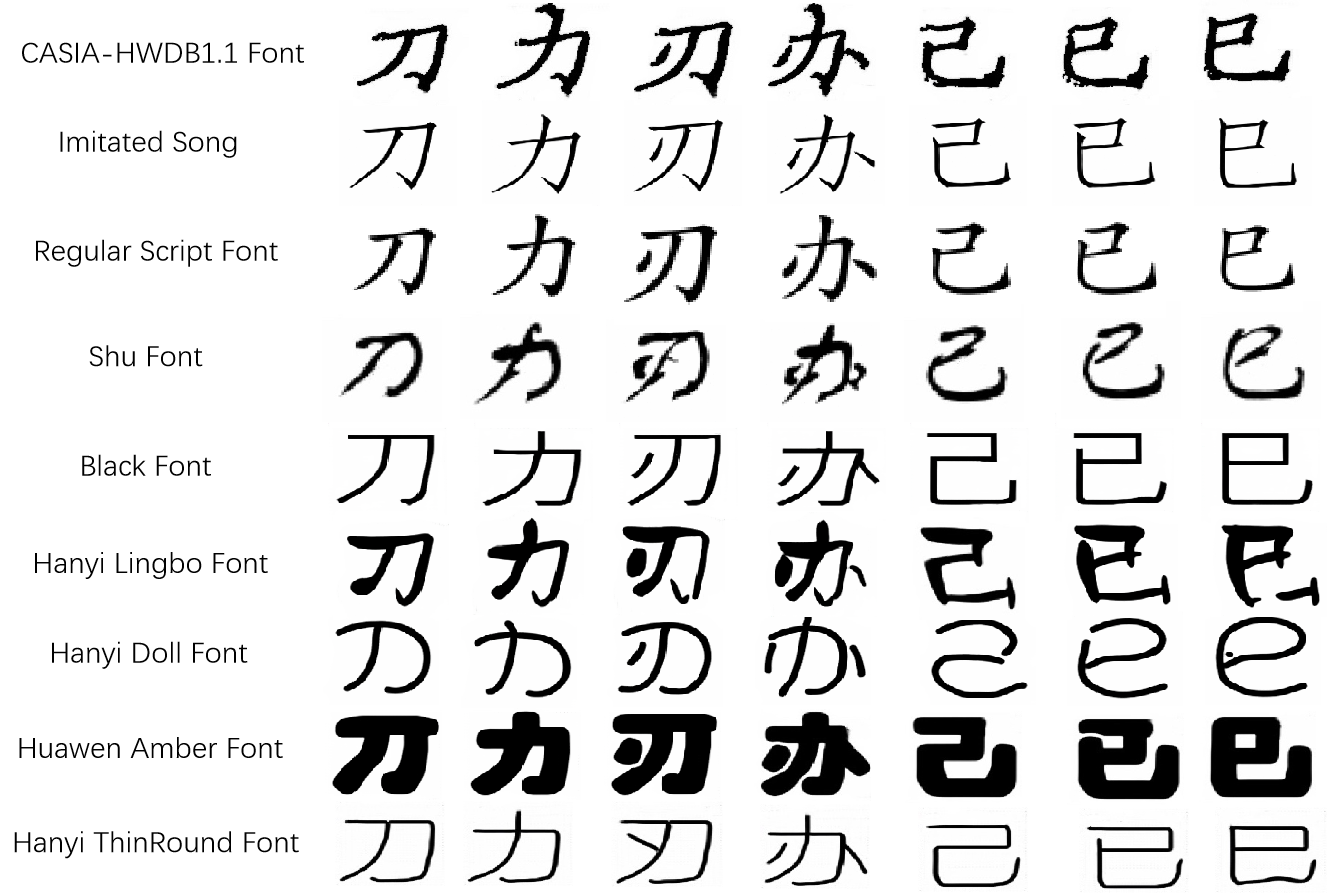}
	\end{minipage}
	\hfill
	\caption{Generated results of StrokeGAN for some very similar Chinese characters with the same stroke encodings.}
	\label{fig:comp-similar-character}
\end{figure}

\begin{figure*}[ht]
	\begin{minipage}[b]{0.99\linewidth}
		\centering
		\includegraphics*[scale=0.45]{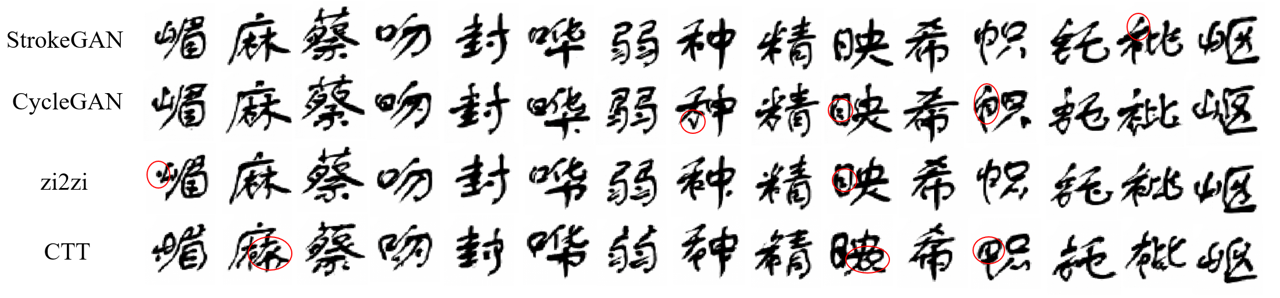}
	\end{minipage}
	\hfill
	\caption{Some examples of Chinese characters generated by StrokeGAN and three existing methods in the generation task from \textit{Regular Script} font to \textit{Shu} font.}
	\label{fig:four-model-result}
\end{figure*}

\subsection{Experiment Results}
\label{sc:exp-results}
Our experiments consist of three parts, where the first two parts were conducted to respectively show the effectiveness and sufficiency of the introduced one-bit stroke encoding,
and the third part was conducted to demonstrate the effectiveness of StrokeGAN via comparing with the state-of-the-art methods including CycleGAN \cite{Chang2018}, zi2zi \cite{Tian2017} and Chinese typography transfer (CTT) method \cite{Chang2017}, where the latter two methods are based on paired training data.

{\bf A. Effectiveness of one-bit stroke encoding.}
In these experiments, we verified the feasibility and effectiveness of our idea that incorporating the stroke information into CycleGAN can preserve the modes of Chinese characters better and thus alleviate the issue of mode collapse. In order to verify this, we implemented nine one-to-one Chinese character style translation experiments as listed in Table \ref{tab:result-number}. For each experiment, we selected one font domain (say \textit{Regular Script}) from these nine different font styles as the source font domain and another font domain (say \textit{Shu}) as the target font domain, and then trained StrokeGAN as well as CycleGAN as the baseline. The performance of the suggested StrokeGAN is presented in Table \ref{tab:result-number}. Some examples of the generated Chinese characters by StrokeGAN are shown in the previous Figure \ref{fig:stgan-results} and Figure \ref{fig:results-4styles}, which demonstrate that StrokeGAN can generate very realistic Chinese characters for all these nine fonts.

From Table \ref{tab:result-number}, StrokeGAN outperforms CycleGAN (without using the stroke encoding) in most of the tasks except the style translation task from \textit{Imitated Song} to \textit{Regular Script}. In terms of the stroke error presented in the last two columns, StrokeGAN consistently improves the performance of CycleGAN \cite{Chang2018}. This shows the feasibility and effectiveness of the suggested StrokeGAN.

Moreover, as demonstrated in the previous Figure \ref{fig:mode-collapse}, CycleGAN sometimes suffers from the issue of mode collapse (particularly, when applied to the generation task from \textit{Black} font to \textit{Shu} font), while the suggested StrokeGAN can significantly alleviate this issue. Besides the issue of mode collapse, as shown in Figure \ref{fig:st-cycle}, CycleGAN also sometimes suffers from another issue, that is, missing some key strokes of Chinese characters generated, which may result in the recognition issues of these characters, while the suggested StrokeGAN can preserve these strokes much better.

{\bf B. Sufficiency of one-bit stroke encoding.} Notice that there are many very similar Chinese characters having the same stroke encodings. In order to show the sufficiency of the introduced one-bit encoding, we tested the performance of StrokeGAN on some very similar Chinese characters,
as shown in Figure \ref{fig:comp-similar-character},
which shows that the generated characters can be well distinguished with well preserved strokes.
This demonstrates that such one-bit encoding way is in general enough to preserve the key modes of Chinese characters.

{\bf C. Comparison with the state-of-the-art methods.}
In this part of experiments, besides CycleGAN \cite{Chang2018}, we compared StrokeGAN with two paired data based methods, i.e., \textit{zi2zi} \cite{Tian2017} and CTT method \cite{Chang2017}.
In order to implement both zi2zi and CTT methods, we manually built up two paired datasets based on the \textit{Regular Script} font and \textit{Shu} font datasets. Since the collection of paired training data is very costly, in these experiments, we only considered the generation task from the \textit{Regular Script} font to \textit{Shu} font for all these four methods, while for the other generation tasks, it can be implemented similarly. The performance of these four methods is presented in Table \ref{4-model-result}, and some examples of generated characters are shown in Figure \ref{fig:four-model-result}. From Table \ref{4-model-result}, the suggested StrokeGAN outperforms all these existing methods in terms of the suggested three evaluation metrics, and also generates the Chinese characters with the highest quality among all these methods. These experiment results demonstrate the effectiveness of the proposed StrokeGAN.

\begin{table}
	
	\caption{Comparison on the performance of the suggested StrokeGAN with existing methods in the generation task from \textit{Regular Script} font to \textit{Shu} font. The first, second and third rows respectively present the content accuracies (\%), recognition accuracies (\%) and stroke errors in the scale $\times 10^{-2}$ of all these four methods.}
	\label{4-model-result}
	\begin{center}
	\footnotesize	
		\begin{tabular}{|c|c|c|c|c|}\hline
			Methods                          & StrokeGAN        & CycleGAN   & zi2zi   & CTT  \\\hline
			Content acc. ($\uparrow$)        & \textbf{90.48}   & 89.56      & 90.12   & 87.92 \\\hline
            Recog. acc.  ($\uparrow$)        & \textbf{90.52}   & 89.36      & 90.04   & 88.01 \\\hline
            Stroke error ($\downarrow$)      & \textbf{5.67}    & 6.79       & 6.36    & 7.52  \\\hline
		\end{tabular}
	\end{center}
\end{table}

%

\section{Conclusion}
\label{sc:conclusion}
This paper proposes an effective Chinese font generation method called \textit{StrokeGAN} by incorporating a one-bit stroke encoding into CycleGAN
to tackle the mode collapse issue. The key intuition of our idea is that the stroke encodings of Chinese characters contain amount of mode information of Chinese characters, unlike the natural images. A new stroke-encoding reconstruction loss was introduced
to enforce a faithful reconstruction of the stroke encoding as accurately as possible and thus preserve the mode information of Chinese characters. Besides the commonly used content accuracy, the crowdsourcing recognition accuracy and stroke error are also introduced to evaluate the performance of our method. The effectiveness of StrokeGAN is demonstrated by a series of Chinese font generation tasks over 9 datasets with different fonts, comparing with CycleGAN and other two existing methods based on the paired data. The experiment results show that StrokeGAN helps preserve the stroke modes of Chinese characters in a better way and generates very realistic characters with higher quality.
Besides Chinese font generation, our idea of the one-bit stroke encoding can be easily adapted to other deep generative models and applied to the font generation related to other languages such as Korean and Japanese.

\section*{Acknowledgment}
The work of Jinshan Zeng is supported in part by the National Natural Science Foundation (NNSF) of China (No.61977038), and Thousand Talents Plan of Jiangxi Province (No. jxsq2019201124). The work of Mingwen Wang is supported in part by NNSF of China (No. 61876074).
The research of Yuan Yao is supported in part by HKRGC 16303817, ITF UIM/390, as well as awards from Tencent AI Lab and Si Family Foundation.
Part of Jinshan Zeng's work was done when he visited at Liu Bie Ju Centre for Mathematical Sciences, City University of Hong Kong.

\bibliography{aaai21}



\begin{table*}[t]
	\caption{Network architecture for the generator of StrokeGAN, which is same to that of CycleGAN in \cite{Zhu2017}.}
	\label{tab:generator-structure}
	\begin{center}
		\footnotesize
		\begin{tabular}{cccc}\hline
			Part                                    & Input  $\rightarrow$ Output Shape        & Layer Information \\\hline
			Input layer                             &$(h,w,3) \rightarrow (h,w,64)$            & BN(CONV-(N64, K7x7, S1, P0), ReLU \\ \hline
			\multirow{2}* {Down-sampling }          &$(h,w,64) \rightarrow (h/2,w/2,128)$      & BN(CONV-(N128, K3x3, S2, P1), ReLU\\
			&$(h/2,w/2,128) \rightarrow (h/4,w/4,256)$ & BN(CONV-(N256, K3x3, S2, P1), ReLU  \\ \hline
			\multirow{2}* {Residual Block}         &$(h/4,w/4,256)\rightarrow (h/4,w/4,256)$  &  (BN(CONV-(N256, K3x3, S1, P1)),  ReLU)   \\
			&$(h/4,w/4,256)\rightarrow (h/4,w/4,256)$  & BN(CONV-(N256, K3x3, S1, P1)  \\\hline
			\multirow{2}* {Residual Block}        	&$(h/4,w/4,256)\rightarrow (h/4,w/4,256)$  & (BN(CONV-(N256, K3x3, S1, P1)), ReLU)   \\
			&$(h/4,w/4,256)\rightarrow (h/4,w/4,256)$  & BN(CONV-(N256, K3x3, S1, P1)  \\\hline
			\multirow{2}* {Residual Block}        	&$(h/4,w/4,256)\rightarrow (h/4,w/4,256)$  & (BN(CONV-(N256, K3x3, S1, P1)), ReLU)   \\
			&$(h/4,w/4,256)\rightarrow (h/4,w/4,256)$  &  BN(CONV-(N256, K3x3, S1, P1)  \\\hline
			\multirow{2}* {Residual Block}        	&$(h/4,w/4,256)\rightarrow (h/4,w/4,256)$  & (BN(CONV-(N256, K3x3, S1, P1)),  ReLU)   \\
			&$(h/4,w/4,256)\rightarrow (h/4,w/4,256)$  &  BN(CONV-(N256, K3x3, S1, P1)  \\\hline
			\multirow{2}* {Residual Block}         &$(h/4,w/4,256)\rightarrow (h/4,w/4,256)$  &  (BN(CONV-(N256, K3x3, S1, P1)),  ReLU)   \\
			&$(h/4,w/4,256)\rightarrow (h/4,w/4,256)$  & BN(CONV-(N256, K3x3, S1, P1)  \\\hline
			\multirow{2}* {Residual Block}       	&$(h/4,w/4,256)\rightarrow (h/4,w/4,256)$  &  (BN(CONV-(N256, K3x3, S1, P1)),  ReLU)   \\
			&$(h/4,w/4,256)\rightarrow (h/4,w/4,256)$  &  BN(CONV-(N256, K3x3, S1, P1)  \\\hline
			\multirow{2}* {Residual Block}         	&$(h/4,w/4,256)\rightarrow (h/4,w/4,256)$  & (BN(CONV-(N256, K3x3, S1, P1)),  ReLU)   \\
			&$(h/4,w/4,256)\rightarrow (h/4,w/4,256)$  &  BN(CONV-(N256, K3x3, S1, P1)  \\\hline
			\multirow{2}* {Residual Block}        	&$(h/4,w/4,256)\rightarrow (h/4,w/4,256)$  & (BN(CONV-(N256, K3x3, S1, P1)),  ReLU)   \\
			&$(h/4,w/4,256)\rightarrow (h/4,w/4,256)$  & BN(CONV-(N256, K3x3, S1, P1)  \\\hline
			\multirow{2}* {Residual Block}      	&$(h/4,w/4,256)\rightarrow (h/4,w/4,256)$  & (BN(CONV-(N256, K3x3, S1, P1)),  ReLU)   \\
			&$(h/4,w/4,256)\rightarrow (h/4,w/4,256)$  &  BN(CONV-(N256, K3x3, S1, P1)   \\ \hline
			\multirow{2}* {Up-sampling}           	&$(h/4,w/4,256)\rightarrow (h/2,w/2,128)$  & BN(DECONV-(N128, K3x3, S2, P1)), ReLU \\
			&$(h/2,w/2,128)\rightarrow (h,w,64)$       & BN(DECONV-(N64, K3x3, S2, P1)), ReLU \\ \hline
			Output layer                            &$(h,w,64)\rightarrow (h,w,3)$             & CONV-(N3,K7xK7,S1,P0),Tanh \\ \hline
		\end{tabular}
	\end{center}
	\label{Network:generator}
\end{table*}

\begin{table*}[h]
	\caption{Network architecture for the discriminator of StrokeGAN, which is similar to that of PatchGAN in \cite{Isola2017} except that the output layers in our model are convolutional layers.}
	\label{tab:discriminator-structure}
	\begin{center}
		\footnotesize
		\begin{tabular}{cccc}\hline
			Layer                     & Input $\rightarrow$ Output Shape                 & Layer Information \\\hline
			Input layer               &$(h,w,3) \rightarrow (h/2,w/2,64)$                & BN(CONV-(N64, K4x4, S2, P1), Leaky ReLU(0.2) \\ \hline
			&$(h/2,w/2,64) \rightarrow (h/4,w/4,128)$          & BN(CONV-(N128, K4x4, S2, P1), Leaky ReLU(0.2)\\
			&$(h/4,w/4,128) \rightarrow (h/8,w/8,256)$         & BN(CONV-(N256, K4x4, S2, P1), Leaky ReLU(0.2) \\
			Hidden layers             &$(h/8,w/8,256) \rightarrow (h/16,w/16,512)$       & BN(CONV-(N512, K4x4, S2, P1), Leaky ReLU(0.2) \\
			&$(h/16,w/16,512) \rightarrow (h/32,w/32,1024)$    & BN(CONV-(N1024, K4x4, S2, P1), Leaky ReLU(0.2) \\
			&$(h/32,w/32,1024) \rightarrow (h/64,w/64,2048)$   & BN(CONV-(N2048, K4x4, S2, P1), Leaky ReLU(0.2) \\ \hline
			Output layer ($D_{src}$)  &$(h/64,w/64,2048)\rightarrow (h/128,w/128,1)$       & CONV-(N1, K4x4, S1, P1)\\
			Output layer ($D_{st}$)  &$(h/64,w/64,2048)\rightarrow (1,1,32)$           & CONV-(N32, K$\frac{h}{64} \times \frac{w}{64}$, S1, P0) \\ \hline
		\end{tabular}
	\end{center}
	\label{Network:discriminator}
\end{table*}


\end{document}